\pdfoutput=1

\documentclass[11pt]{article}

\usepackage[]{acl}

\usepackage{times}
\usepackage{latexsym}

\usepackage[T1]{fontenc}

\usepackage[utf8]{inputenc}

\usepackage{microtype}
\usepackage{array}
\usepackage{booktabs}
\usepackage{multirow}
\usepackage{amsmath}
\usepackage{graphicx}
\usepackage{marginnote}
\usepackage{enumitem}
\usepackage{xurl}
\usepackage[colorinlistoftodos,prependcaption,textsize=tiny]{todonotes}

\definecolor{bluepigment}{rgb}{0.2, 0.2, 0.6}

\usepackage{xparse}

\urldef{\codeurl}\url{https://www.amazon.science/publications/enhanced-knowledge-selection-for-grounded-dialogues-via-document-semantic-graphs }

\ifx\nocomment\undefined
\NewDocumentCommand{\di}
{ mO{} }{\textcolor{green}{\textsuperscript{\textit{Di}}\textsf{\textbf{\small[#1]}}}}
\NewDocumentCommand{\heng}
{ mO{} }{\textcolor{red}{\textsuperscript{\textit{Heng}}\textsf{\textbf{\small[#1]}}}}
\NewDocumentCommand{\mbc}
{ mO{} }{\textcolor{cyan}{\textsuperscript{\textit{Mohit}}\textsf{\textbf{\small[#1]}}}}
\NewDocumentCommand{\zoey}
{ mO{} }{\textcolor{orange}{\textsuperscript{\textit{Zoey}}\textsf{\textbf{\small[#1]}}}}
\NewDocumentCommand{\yang}
{ mO{} }{\textcolor{purple}{\textsuperscript{\textit{Yang}}\textsf{\textbf{\small[#1]}}}}
\else 
\NewDocumentCommand{\heng}{ mO{} }{\textcolor{red}{}}
\NewDocumentCommand{\zoey}{ mO{} }{\textcolor{red}{}}
\NewDocumentCommand{\yang}{ mO{} }{\textcolor{red}{}}
\NewDocumentCommand{\mbc}{ mO{} }{\textcolor{red}{}}
\NewDocumentCommand{\di}{ mO{} }{\textcolor{red}{}}
\fi

\title{Enhanced Knowledge Selection for Grounded Dialogues via Document Semantic Graphs}

       \author{ Sha Li$^1$\thanks{Work done as an intern at Amazon Alexa AI.} , Mahdi Namazifar$^2$, Di Jin$^2$, Mohit Bansal$^{2}$, Heng Ji$^{2}$,
       \\
       \textbf{Yang Liu}$^2$, \textbf{Dilek Hakkani-Tur}$^2$
       \\
       $^1$University of Illinois at Urbana-Champaign, $^2$ Amazon Alexa AI \\
       \tt{shal2@illinois.edu} \\
       \tt{\{mahdinam, djinamzn, mobansal, jihj,}\\
       \tt{yangliud, hakkanit\}@amazon.com}
       }

\begin{document}
\maketitle
\begin{abstract}

Providing conversation models with background knowledge has been shown to make open-domain dialogues more informative and engaging. 
Existing models treat knowledge selection as a sentence ranking or classification problem where each sentence is handled individually, ignoring the internal semantic connection among sentences in background document.
In this work, we propose to
automatically convert the background knowledge documents into \textit{document semantic graphs} and then perform knowledge selection over such graphs.
Our document semantic graphs preserve sentence-level information through the use of sentence nodes and provide concept connections between sentences. 
We apply multi-task learning for sentence-level knowledge selection and concept-level knowledge selection jointly, and show that it improves sentence-level selection.
Our experiments show that our semantic graph based knowledge selection improves over sentence selection baselines
for both the knowledge selection task and the end-to-end response generation task on HollE~\cite{moghe-etal-2018-HollE} and improves generalization on unseen topics in WoW~\cite{Dinan2019WizardOW}.\footnote{See \codeurl  for an updated paper with information about code and resources.}

\end{abstract}

\section{Introduction}
\begin{figure}
    \centering
    \includegraphics[width=.8\linewidth]{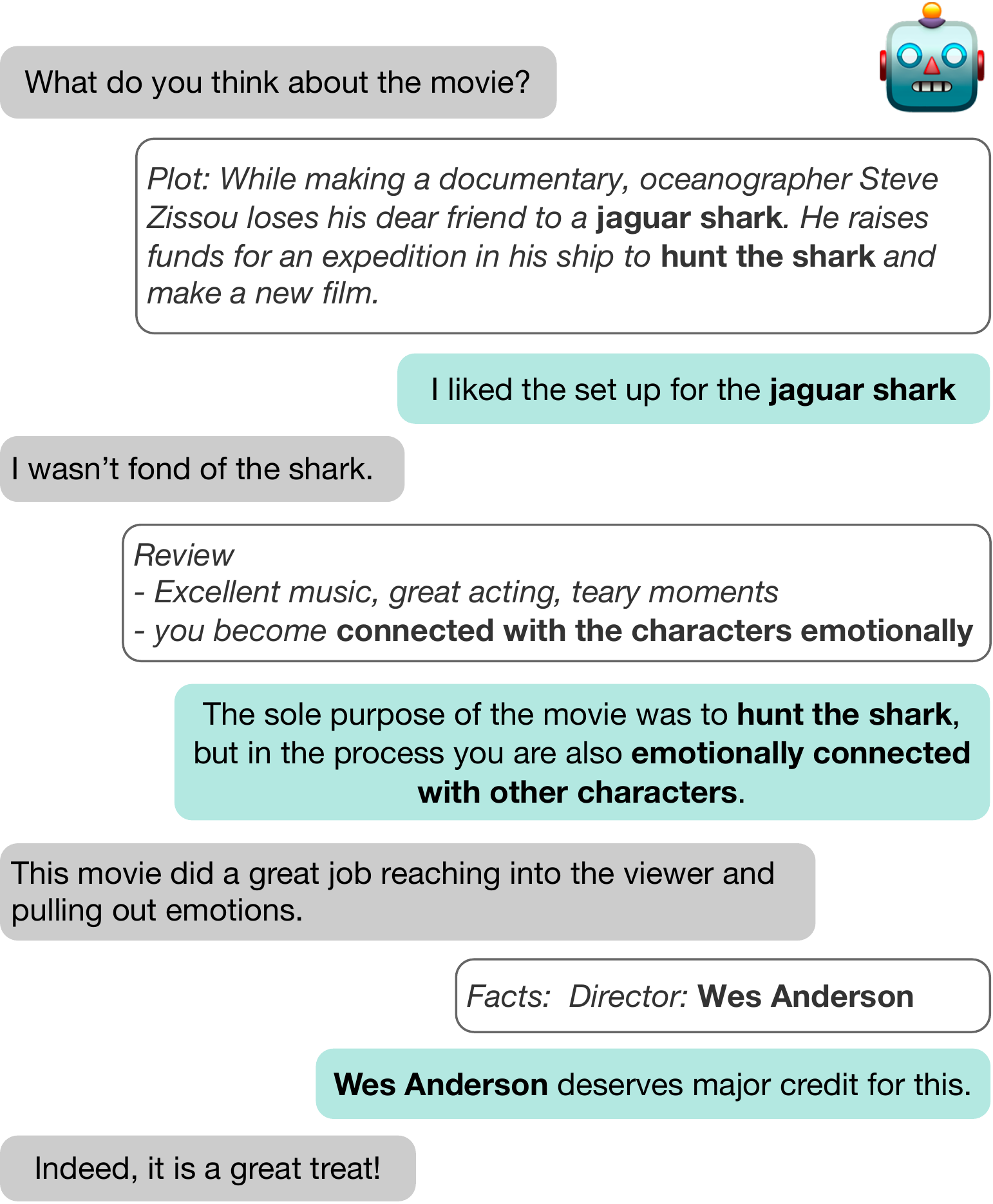}
    \caption{An example of knowledge-grounded dialog. 
    Semantic connections between sentences improve coherence and not imposing knowledge boundaries allows the system to utilize multiple knowledge snippets. The used knowledge is in bold.
    *The jaguar shark is a character.
    }
    \label{fig:grounded_dialog}
\end{figure}

Natural language generation models have seen great success in their ability to hold open-domain dialogues without the need for manual injection of domain knowledge.
However,
such models often degenerate to uninteresting and repetitive responses~\cite{Holtzman2020TextDegeneration}, or hallucinate false knowledge~\cite{Roller2021RecipesFB,Shuster2021RetrievalAR}. 
To avoid such phenomena, one solution is to provide the conversation model with relevant knowledge to guide the response generation~\cite{parthasarathi-pineau-2018-extending,Ghazvininejad2018AKN,Dinan2019WizardOW}.  Figure~\ref{fig:grounded_dialog} illustrates such knowledge grounded generation.

Relevant knowledge is often presented in the form of documents~\cite{moghe-etal-2018-HollE,Zhou2018CMUDocumentGrounded, Ghazvininejad2018AKN, Dinan2019WizardOW, Gopalakrishnan2019TopicalChatTK} and the task of identifying the appropriate knowledge snippet for each turn is formulated as a sentence classification or ranking task~\cite{Dinan2019WizardOW}. Although more advanced methods have been proposed by modeling knowledge as a latent variable~\cite{Kim2020SequentialLK, chen-etal-2020-bridging}, or tracking topic shift~\cite{Meng2021InitiativeAwareSL}, they abide by the setting of sentence-level selection. This setting has two inherent drawbacks: (1) it ignores the \textit{semantic connections} between sentences and (2) it imposes an artificial constraint over the \textit{knowledge boundary}.

\begin{figure}[t]
    \centering
    \includegraphics[width=\linewidth]{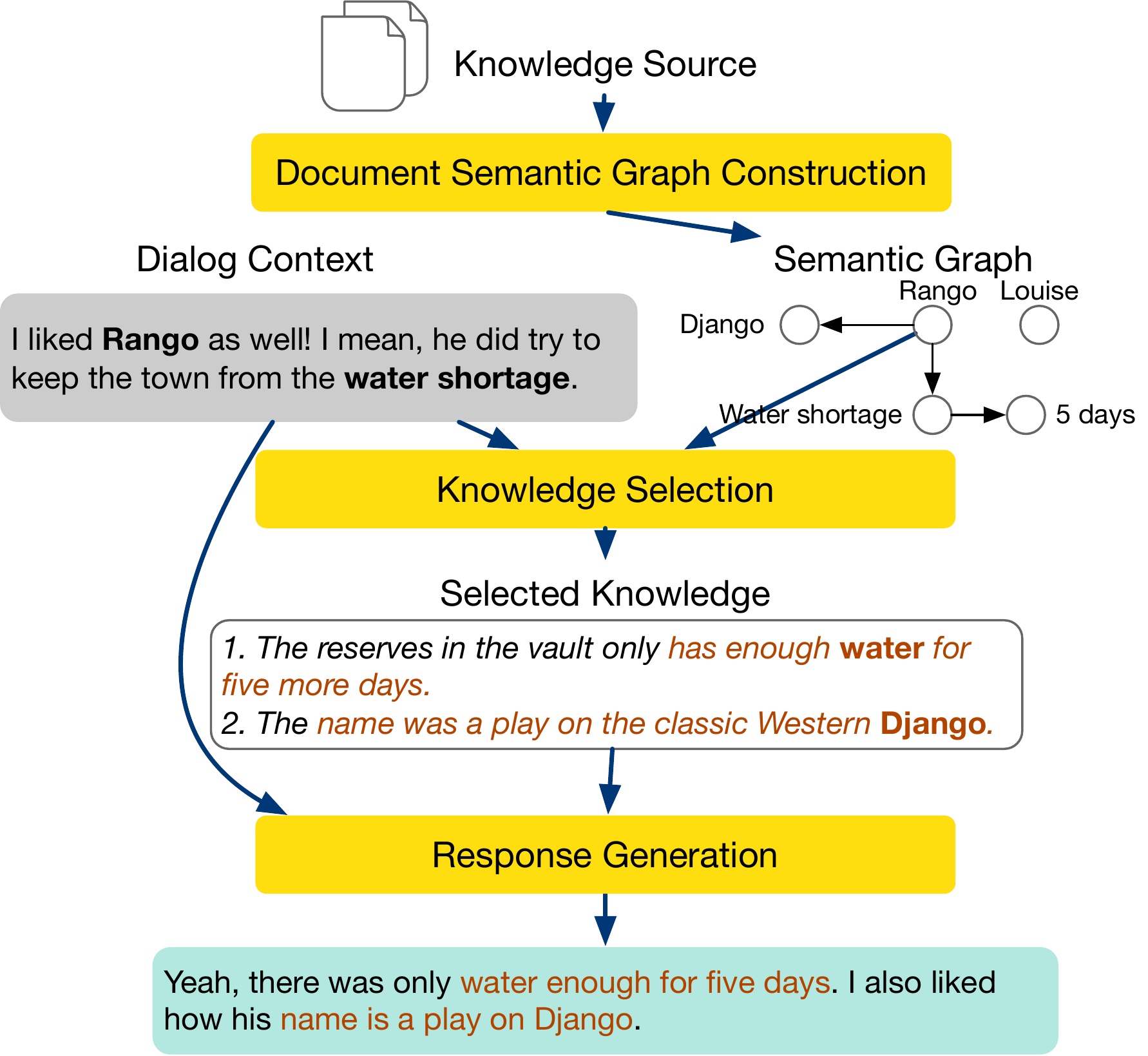}
    \caption{The pipeline for generating responses based on a given knowledge source. }
    \label{fig:pipeline}
\end{figure}

A document is not simply a bag of sentences, in fact, it is the underlying semantic connections and structures that make the composition of sentences meaningful. Two examples of such connections are coreference links  and predicate-argument structures.\footnote{Another example would be discourse relations between sentences, which we do not explore here.} These connections are vital to the understanding of the document and also beneficial to knowledge selection. In many cases, we can draw information from multiple sentences to create the response, breaking the \textit{knowledge boundary}. 
For instance, in Figure \ref{fig:pipeline}, the connections among the character ``Rango'', the plot point ``water shortage" and the name ``Django" help us generate a response with a smooth topic transition.

A related line of work~\cite{liu-etal-2018-knowledge,moon-etal-2019-opendialkg, Xu2020KnowledgeGG,Young2018AugmentingED, Zhou2018CommonsenseKA} that seemingly overcomes the aforementioned issues is knowledge selection from existing knowledge graphs (KGs) such as Wikidata~\cite{Vrandei2014WikidataAF}, DBpedia~\cite{lehmann2015dbpedia}, and ConceptNet~\cite{Speer2017ConceptNet5A}. If the character ``Rango''  were in the KG, it would have been represented as an entity node and be connected to respective events. On the KG, we are also free to select as many concepts as needed, without being restricted to a single sentence as the source. However, KGs are known to have limited coverage of real world entities, let alone emerging entities in works of fiction such as books and movies~\cite{Razniewski2016ButWD}.

Hence, to bridge these two worlds of sentence-based knowledge selection and KG-based knowledge selection, we introduce knowledge selection using \textit{document semantic graphs}.
These graphs are automatically constructed from documents, aiming to preserve the document content while enhancing the document representation with semantic connections.
To create such a document semantic graph, we first obtain the Abstract Meaning Representation (AMR)~\cite{Banarescu2013AMR} for each sentence. AMR detects entities, captures predicate-argument structures, and provides a layer of abstraction from words to concepts.\footnote{In AMR, every node is a concept. This includes events, objects, attributes, etc.} 
Compared to existing knowledge base construction methods, AMR covers a wide range of  relations and fine-grained semantic roles, and can fully reflect the semantics of the source text. 
Since AMR graphs only represent single sentences, we utilize coreference resolution tools to detect coreferential entity nodes and merge them to build graphs for documents. On top of this content representation, we also add sentence nodes and passage nodes to reflect the structure of the document.
This allows for traversal across the graph by narrative order or concept association.

Given the document semantic graph, knowledge selection can be seen as identifying relevant nodes on the graph,  sentence nodes or concept nodes.
As knowledge selection in dialog models is conditioned on the dialog context, for each dialog turn, we create a \textit{dialog-aware graph} derived from the document graph. It contains context nodes representing contextualized versions of the sentence and concept nodes. 
We design an edge-aware graph neural network model to propagate information along the dialog-aware graph and finally score the context nodes (or concept nodes) on their relevance to the dialog turn (as shown in Figure \ref{fig:joint_graph}).

We validate our model on two widely used datasets HollE~\cite{moghe-etal-2018-HollE} and Wizard of Wikipedia~\cite{Dinan2019WizardOW} by constructing a semantic graph from relevant background documents\footnote{On WoW we use the passages retrieved from the first turn for graph construction. Our method will need to be extended to online graph construction to support per-turn retrieval of documents.} for each dialog. The use of document graphs improves both knowledge selection and response generation quality on HollE and boosts generalization to unseen topics for WoW.
From our ablation tests we find that in terms of the graph structure, the key component is the use of coreference edges that stitches sentences together.

Our contributions include: 
(1) We propose to perform knowledge selection from document semantic graphs that are automatically constructed from source documents and can reflect the implicit semantic connections between sentences without being limited to a pre-defined set of entities and relations as KGs do. 
Our approach bridges the gap between sentence-based knowledge selection and KG-based knowledge selection.
(2) We show that joint selection over sentences and concepts can 
model more complex relations between sentences and boost sentence selection performance.
(3) We build a pipeline for converting documents (document collections) into semantic graphs through the use of AMR parsing and coreference. We hope that our tool can facilitate future work on graph-based representations of documents.

\section{Method}
We show an overview of our knowledge-grounded dialog system in Figure \ref{fig:pipeline}. The system consists of three modules, namely semantic graph construction, knowledge selection and response generation.

\subsection{Document Semantic Graph Construction}

\begin{figure}[t]
    \centering
    \includegraphics[width=\linewidth]{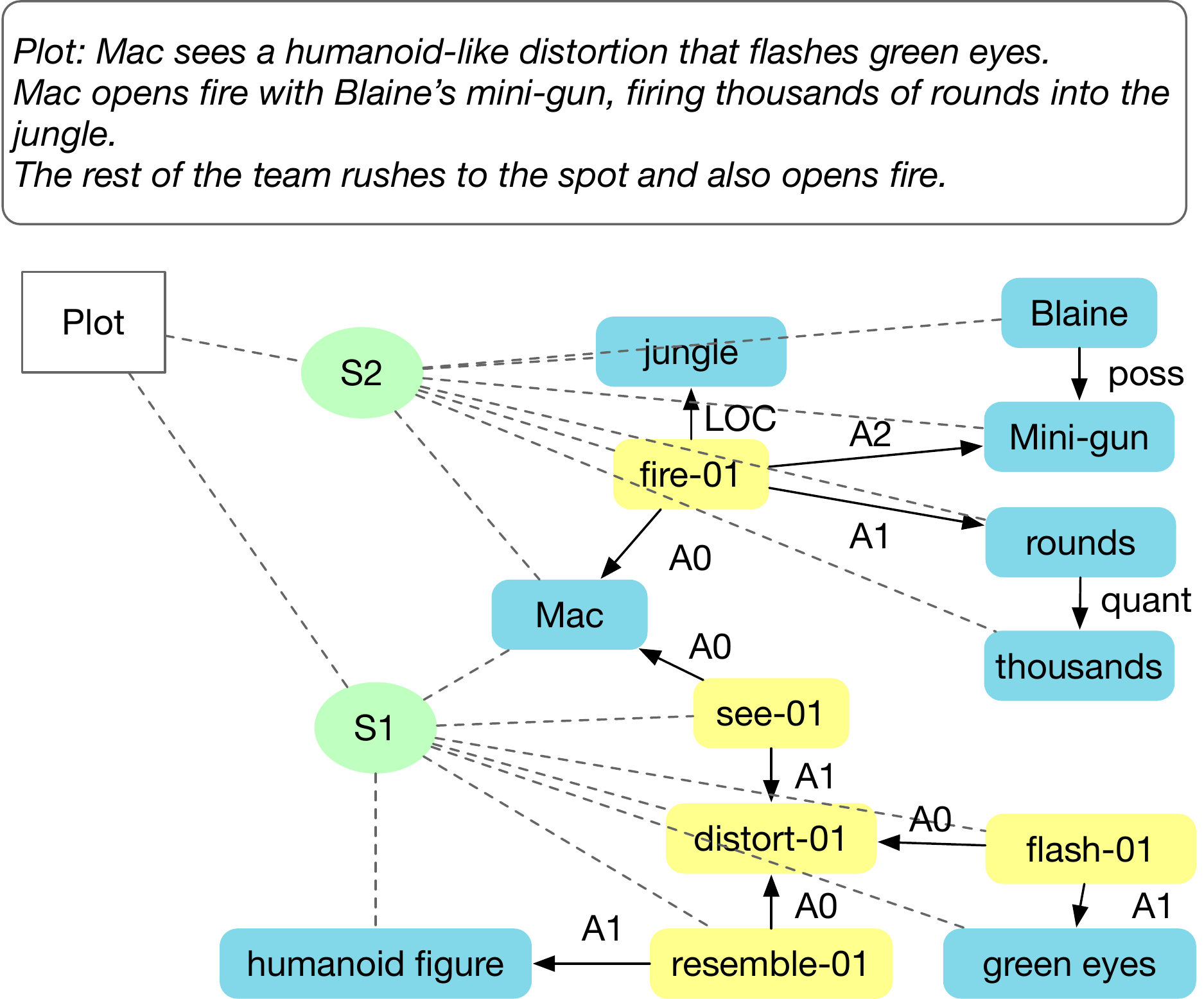}
    \caption{Part of the document semantic graph for the shown plot. The graph includes the source node (white rectangle), the sentence nodes (green circles), and the concept nodes (yellow and blue rectangles). Directional edges with labels (e.g., A0, A1) are from AMR parsing, dotted edges are from the document structure.}
    \label{fig:kg}
\end{figure}

We first process the sentences in the background knowledge documents using the Stack Transformer AMR parser~\cite{fernandez-astudillo-etal-2020-transition} to obtain sentence-level AMR graphs.
Based on the AMR output, we consider all of the concepts %
that serve as the core roles (agent, recipient, instrument etc.) for a predicate as mention candidates.
Then, we run a document-level entity coreference resolution system~\cite{wen-etal-2021-resin} to resolve coreference links between such mentions. 
When joining sentence-level AMR graphs to form the document graph, entity mentions %
that are predicted to be coreferential are merged into one node, and we keep the longest %
mention as the node's canonical name. We show an example of our constructed document semantic graph in Figure \ref{fig:kg}.

On top of this content representation, we also add additional nodes to represent documents (or passages) and sentences. 
A document (or passage) node is linked to sentence nodes that are from this document (or passage).
Each sentence node is directly connected to all the concept nodes that originate from that sentence. 
In addition, we add edges between neighboring sentences following the narrative order in the document.

Since each node is grounded in text, in order to create embeddings for the document semantic graph,
we initialize the embedding of each node with their contextual embeddings from a frozen pretrained language model RoBERTa~\cite{Liu2019RoBERTaAR}.
For sentence nodes, we use the embedding of the \texttt{[CLS]} token. For concept nodes, we average the embeddings of the tokens in the span.
Note that the document semantic graphs can be created offline and indexed by topics to be used at knowledge selection inference time.

\begin{figure*}
    \centering
    \includegraphics[width=\linewidth]{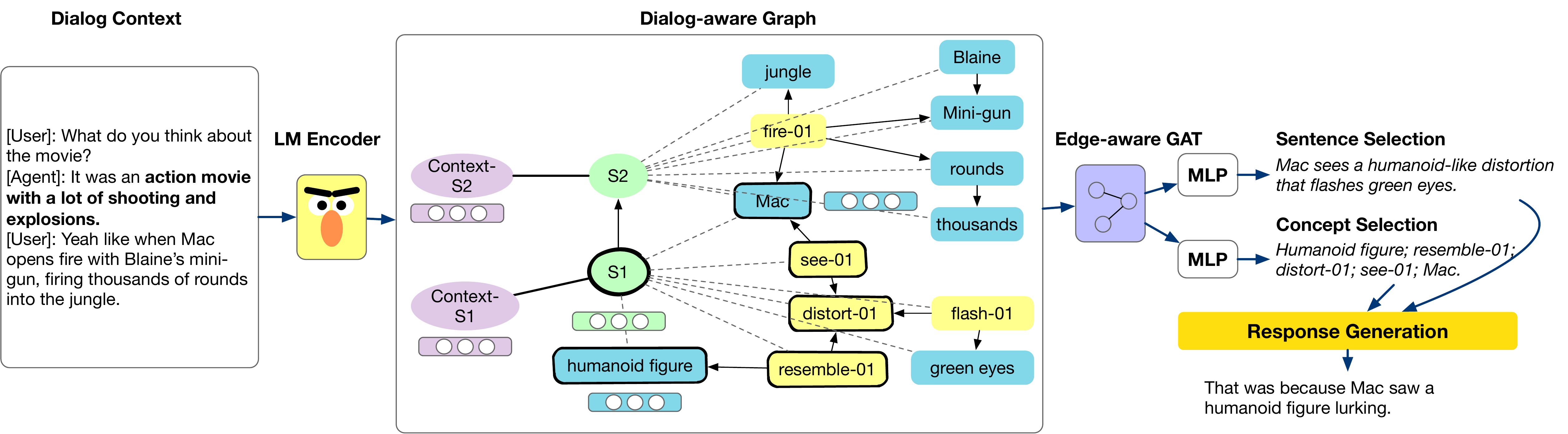}
    \caption{The knowledge selection model. We encode the dialog context using a pretrained language model and represent the dialog context along with each candidate sentence as a context node. We then use an edge-aware graph attention network to encode the dialog-aware graph. Finally, we classify each node on the graph to be relevant or not based on the learned node embedding, effectively performing both sentence selection and concept selection. The selected nodes are outlined in black. 
    }
    \label{fig:joint_graph}
\end{figure*}

\subsection{Knowledge Selection}
The task of knowledge selection is to identify relevant knowledge snippets that can be used to produce an appropriate and informative response for each turn. Since our document semantic graph is based on the background knowledge source alone, we first create a \textit{dialog-aware graph} that is conditioned on the given dialog turn.
We then encode the dialog-aware graph by an edge-aware graph attention network and predict relevance scores for sentences and concepts as shown in Figure \ref{fig:joint_graph}.

\paragraph{Dialog-Aware Graph.}

The dialog-aware graph is a copy of the document semantic graph with additional context nodes ($c$), each representing a dialog-contextualized knowledge sentence.
For each candidate knowledge sentence $s_i$, to obtain the embedding $h_c$ of the context node $c_i$, we encode the dialog context $x$ and the candidate knowledge sentence $s_i$ through a pretrained language model $f_{\texttt{LM}}$.
We define the dialog context as the most recent two turns in the dialog history.
\vspace{-0.05in}
\begin{equation}
    h_{c_i} = \text{Pooling} \left(f_{\texttt{LM}}([s_i;x ])\right)
\end{equation}
For the pooling operation, we simply take the first token (namely the \texttt{[CLS]} token) as the representation for the sequence. 
Since we want to enable message passing between the context node and the rest of the graph, we add an edge between the context node $c_i$ and the sentence node $s_i$.

\paragraph{Edge-Aware Graph Attention Network.}
At this point, although our dialog-aware graph captures both the dialog context and the knowledge source, there is no interaction between the two. To this end, we apply an edge-aware graph attention network (EGAT) model to allow information to be propagated along the graph.
Note that our dialog-aware graph is a heterogeneous network with multiple node types and edge types.
To capture the semantics of the node and edge types, we use an extension of the graph attention network~\cite{Velickovic2018GAT} that includes edge type embeddings $h_{T(e)}$ and node type embeddings $h_{T(v)}$~\cite{yasunaga-etal-2021-qa}. These embeddings are learnt along with the model parameters and are used to compute the vector ``message'' that is passed along edges.

In general, a graph neural network consists of $L$ layers with shared parameters. We denote the initial embeddings for each node as $h^0$. Each layer $l$ involves a round of nodes sending out ``messages'' to their neighbors and then aggregating the received ``messages'' to update their own embeddings from $h^l$ to $h^{l+1}$.
Consider a pair of nodes $s$ and $t$ with embeddings $h_s$ and $h_t$ respectively, the message $m_{s \rightarrow t}$ that is passed from $s$ to $t$ through edge $e$ is computed as the sum of the edge-aware message and the node-aware message, where $W_v$ and $W_e$ are projection matrices.
\begin{equation}
     m_{s\rightarrow t} = W_v( [h^l_s; h_{T(v)}])  + W_e h_{T(e)}
\end{equation}
Then we compute the attention weight $\alpha_{s \rightarrow t}$ from node $s$ to node $t$ as:
\begin{equation}
\small
\begin{aligned}
    q_s &= W_q([h^l_s; h_{T(s)}]) \\
    k_t &= W_k([h^l_t; h_{T(t)}; h_{T(e)}]) \\
    \alpha_{s\rightarrow t} & = \text{Softmax}_{s \in \mathcal{N}_t} \left(\frac{q_s^T k_t}{\sqrt{D}}\right)
\end{aligned} 
\end{equation}
Here $W_q$ and $W_k$ are learnt projection matrices and $D$ is the embedding dimension of $h_s$. $\mathcal{N}_t$ is the neighbor node set of node $t$.
Finally, the messages from the surrounding neighbors are aggregated to compute the updated node embedding $h^{l+1}_t$.
\vspace{-0.05in}
\begin{equation}
\nonumber
\begin{aligned}
\small 
    h^{l+1}_t & = \text{GELU} \left( \text{MLP}(\sum_{s \in \mathcal{N}(t)} \alpha_{s\rightarrow t} m_{s \rightarrow t}) + h_t^l \right)
\end{aligned}
\label{eq:graph-prop}
\end{equation}
After $L$ layers, we obtain embeddings for our context nodes $h_c^L$, sentence nodes $h_s^L$ and concept nodes $h_n^L$.

\paragraph{Knowledge Selection Training.}

For each context node $c$ that represents a pair of the knowledge sentence and dialog context, we compute their relevance score as
\vspace{-0.05in}
\begin{equation}
    \text{score}(c) = \text{MLP}([h_c^L; h_c^0]) 
\end{equation}
For each concept node $n$, we compute its relevance score as
\vspace{-0.05in}
\begin{equation}
    \text{score}(n) = \sigma \left( \text{MLP} (h_n^L) \right)  
\end{equation}
where $\sigma$ is the sigmoid function.

Each context node $c$ needs to be encoded with the language model $f_{LM}$, but we are unable to fit all context nodes into memory.\footnote{On average, we have 60 knowledge sentences per turn.}
Hence, during training, we randomly sample $k$ negatives for each positive knowledge sentence and compute cross-entropy loss over the samples.
\vspace{-0.05in}
\begin{equation}
    \mathcal{L}_c = - \log \frac{\exp \left( \text{score}(c^+) \right) }{\exp_{c \in \{c^+\} \cup C^-} \left(\text{score}(c) \right)
    }
\end{equation}

For concept nodes, we treat knowledge selection as a binary classification problem and compute the binary cross entropy loss.

\begin{equation}
    \mathcal{L}_n = - \frac{1}{N} \sum_{n \in G} r_n \log \text{score}(n)
\end{equation}
Here $r_n \in \{0, 1\}$ is the relevance label for the vertex $n$, and $N$ is the total number of concept nodes. When the dataset does not directly provide concept-level labels for training, we derive them from the ground truth knowledge snippet by assigning any concept that is mentioned in the snippet with a relevant label $r_n =1$.
The overall loss is the weighted sum of the above sentence-level and the concept-level loss:
\begin{equation}
    \mathcal{L} = \mathcal{L}_c + \beta \mathcal{L}_n 
\end{equation}

During inference, we compute $\text{score}(c)$ for all knowledge sentence candidates and take the highest scored sentence for knowledge grounded response generation.

\subsection{Response Generation}
We fine-tune a left-to-right language model GPT2~\cite{Radford2019LanguageMA} to perform response generation given the dialog context $x$ and the chosen knowledge snippet $\hat s$.
\vspace{-0.05in}
\begin{equation}
    y = \text{GPT2}([\hat s; x]) 
\end{equation}
During training we use teacher-forcing and use the ground truth knowledge snippet. This response generation model is independent from the knowledge selection model and trained with negative log-likelihood loss.

\section{Experiments}
\subsection{Datasets}
\begin{table}[t]
    \centering
    \resizebox{\linewidth}{!}{ 
    \begin{tabular}{l l c c c}
    \toprule 
        Dataset & & Train & Dev & Test \\
    \midrule 
    \multirow{2}{3em}{HollE}   & Dialogs &   7,228 & 930 & 913 \\
    & \# turns & 34,486 & 4,388 & 4,318 \\
    \multirow{2}{3em}{WoW} & Dialogs & 17,629 & 941/936 & 924/952 \\
    & \# turns & 22,715 & 3257/3085 & 3104/3298  \\
    \bottomrule
    \end{tabular}}
    \caption{Dataset statistics for WoW and HollE. For WoW, the first column is the seen split and the second column is the unseen split.}
    \label{tab:dataset}
\end{table}

We evaluate our model on two publicly available datasets: Wizard of Wikipedia~\cite{Dinan2019WizardOW} and Holl-E~\cite{moghe-etal-2018-HollE}. Both datasets are in English.

Wizard of Wikipedia (WoW) is an open-domain dialog dataset, spanning multiple topics including famous people, works of art, hobbies, etc.
The test set in WoW consists of two splits that are named ``Test Seen'' and ``Test Unseen'' based on the overlap of topics with the training set.
In order to build our document graph, we use the selected topic passage and the passages retrieved in the first turn as background knowledge. 

Holl-E is a movie domain dialog dataset. Each dialog discusses one movie, and the background knowledge includes the plot, reviews, comments and a fact table. 
Holl-E additionally provides multiple references for the test set so we report performance for both single  and multiple references.

\subsection{Implementation Details}
\paragraph{Knowledge Selection.} 
We only use the turns that utilize knowledge for training and prediction.
To map the ground truth knowledge to a set of concept nodes, we choose all nodes with mention offsets contained within the span. We acquire the sentence-level labels following \cite{Kim2020SequentialLK}.

We use Roberta-base~\cite{Liu2019RoBERTaAR} as the language model $f_{LM}$.
We set $k=5$ for negative sampling.
The EGAT model is trained with 200 hidden dimensions and 2 layers. Edge features and node features are represented with 20 dimensional vectors.
We train our model with a batch size of 16 and learning rate $3e^{-5}$ for 3 epochs.

\paragraph{Response Generation.} 

Our response generation model is based on GPT2~\cite{Radford2019LanguageMA} and is further fine-tuned with a batch size of 16 and learning rate of $3e{^-5}$. We truncate the dialog context to 128 tokens.
During inference, we adopt top-k and top-p sampling with $k=20$ and $p=0.95$. The maximum generation length is limited to 286 tokens, including the input tokens.

\subsection{Baselines}
For knowledge selection, we also implemented the following two baseline methods: 
\begin{itemize}[leftmargin=*]
  \setlength\itemsep{0em}
    \item Roberta Ranking. We use a cross-encoder based on Roberta to represent the dialog context and the knowledge candidate, and a classification layer on top of it.
    \item Graph Paths. This model is built on top of the previous model. The graph paths are from the document semantic graph and obtained by breadth-first traversal starting at the candidate context node. In order to utilize the graph paths, we linearize it into tuples of (subject, predicate, object) or (modifier, subject)  according to the AMR edge label and concatenate it with the candidate sentence.
\end{itemize}

For the end-to-end pipeline, we use the GPT2 response generation with our knowledge selection module and the two methods above.
In addition, we compare against the following previous methods: 
\begin{itemize}[leftmargin=*]
  \setlength\itemsep{0em}
    \item Transformer MemNet~\cite{Dinan2019WizardOW} is the combination of a Transformer memory network for knowledge selection and another Transformer decoder for generation. 
    \item E2E BERT is a variant of the previous model using BERT~\cite{devlin-etal-2019-bert}.
    \item Sequential Knowledge Transformer (SKT)~ \cite{Kim2020SequentialLK} models knowledge as a latent variable and considers the posterior distribution of knowledge given the response. %
    \item SKT+PIPM+KDBTS~\cite{chen-etal-2020-bridging} is an improvement upon SKT with an additional Posterior Information Prediction Module (PIPM) and trained with knowledge distillation.
    \item Mixed Initiative Knowledge Selection (MIKe)~\cite{Meng2021InitiativeAwareSL} uses two knowledge selection modules to capture user-driven turns and system-driven turns respectively. 
   
\end{itemize}

\subsection{Evaluation Metrics}
\paragraph{Knowledge Selection.}
To compare with previous methods, we use Accuracy, or Precision@1 as the main metric for evaluating knowledge selection.
Additionally, we compute sentence ranking metrics, namely the mean average precision (MAP) and mean reciprocal rank (MRR)\footnote{\url{https://github.com/usnistgov/trec_eval}} for more fine-grained analysis of knowledge selection quality.

\paragraph{Response Generation.} 

For automatic evaluation of responses, we use ROUGE-1, ROUGE-2 and ROUGE-L metrics~\cite{lin-2004-rouge}.\footnote{We use the \texttt{torchmetrics} package, which follows \texttt{rouge-score} package Python ROUGE implementation.} 
As our response generation model is trained with gold-standard knowledge, we only report perplexity scores when using gold-standard knowledge, as a measure for the quality of the response generator alone.

For our human evaluation, we randomly sample 200 turns from the output of MIKe~\cite{Meng2021InitiativeAwareSL}, our ranking model and our graph-based model.
Annotators are asked to select which system's response is the best among the three (allowing for ties), and which system's knowledge is the most relevant. In addition, annotators score each response based on whether it is appropriate, knowledgeable and engaging on a scale of 1-4. Our annotators agreed with each other 54.2\% on a single system and 91.7\% when accounting for ties.  The Krippendorff's alpha score for the normalized appropriate/knowledgeable/engaging scores is 0.537/0.634/0.470.

\begin{table}[t]
    \centering
    \footnotesize
    \begin{tabular}{l|c  c | c c c }
    \toprule 
       Model  & \multicolumn{2}{c | }{Single Reference} & \multicolumn{3} {c} {Multiple Reference} \\
      \midrule 
      & MAP  &  Acc  & MAP & MRR &  Acc \\
      \midrule 
   Ranking & 0.493 & 34.3 & 0.527 & 0.526 & 45.3 \\
   Graph Paths & 0.497 & 35.0  & 0.527 & 0.579 & 45.8 \\
   Ours & 0.513  & 37.7** & 0.514 & 0.580 & 46.1 \\
    \bottomrule 
    \end{tabular}
    \caption{Knowledge selection results on the HollE dataset. For single references, MRR is the same as MAP. Acc is reported in percentage\%. ** indicates significance compared to the second best model with $p< 0.005$ under the paired t-test.  }
    \label{tab:ks_holle}
\end{table}

\begin{table}[t]
    \centering
    \footnotesize
    \begin{tabular}{l|c  c | c  c }
    \toprule 
       Model  & \multicolumn{2}{c | }{Test Seen} & \multicolumn{2} {c} {Test Unseen} \\
      \midrule 
      & MAP  &  Acc  & MAP &  Acc \\
      \midrule 
   Ranking & 0.472 & 30.1 & 0.436 & 26.3 \\
   Graph Paths & 0.469 & 29.5 & 0.436 & 26.4 \\
   Ours & 0.469 & 29.4 & 0.486 & 30.8** \\
    \bottomrule 
    \end{tabular}
    \caption{Knowledge selection results on WoW using the topic passage and passages retrieved at the first turn. Acc is reported in percentage\%.   ** indicates significance compared to the second best model with $p< 0.005$ under the paired t-test. 
    }
    \label{tab:ks_wow}
\end{table}

\begin{table*}[t]
    \centering
    \resizebox{0.8\textwidth}{!}{
    \begin{tabular}{l|c c c  | c c c  }
    \toprule 
   Model & \multicolumn{3}{c|}{Single Reference}& \multicolumn{3}{c} {Multiple Reference}  \\
        & R1 & R2 & RL & R1 & R2 & RL  \\
        \midrule 
       Transformer MemNet~\cite{Dinan2019WizardOW} & 20.1 & 10.3  & -  & 24.3 & 12.8 & -  \\
       E2E BERT $\dagger$ & 25.9 & 18.3 & -  & 31.1 & 22.7 & -  \\
       SKT~\cite{Kim2020SequentialLK} & 29.8 & 23.1 & -  & 36.5 & 29.7 &  -  \\
       SKT+PIPM+KDBTS~\cite{chen-etal-2020-bridging} & 30.8 & 23.9 & -   & 37.7 & 30.7 & -  \\ 
       MIKe~\cite{Meng2021InitiativeAwareSL} & 37.78 & 25.31 & 32.82 & 44.06 & 31.92 & 38.91 \\
       GPT2 + Ranking & 40.22 & 31.78 & 38.73  &  47.53 & 39.31 & 45.89 \\
       GPT2 + Graph Paths & 40.76 & 32.32 & 39.12 &47.71 & 39.33 & 45.90 \\
       GPT2 + Graph Selection & 42.49 & 34.37 & 41.01 &47.89 &39.58 &46.25\\
       \midrule 
       GPT2 + Gold knowledge & 75.92 & 72.82 & 75.37  & 75.92 & 72.82 & 75.37 \\
       \bottomrule 
    \end{tabular}}
    \caption{Response generation results ROUGE-1 (R1), ROUGE-2 (R2), ROUGE-L (RL) on HollE. $\dagger$ results taken from \cite{Kim2020SequentialLK}. Other results with citations are taken from their respective papers.}
    \label{tab:exp_holle}
\end{table*}

\begin{table}[t]
    \centering
    \footnotesize
    \begin{tabular}{l|c | c c c }
    \toprule 
    Model  &  Preferred & Approp. & Know. & Engaging \\
    \midrule 
    Ours  &  69\%  & 3.54 & 3.42 & 3.32 \\
     Ranking & 56\% & 3.47 & 3.39 & 3.28 \\
     MIKe & 34.5\% & 2.88 & 3.02 & 2.82 \\
     \bottomrule 
    \end{tabular}
    \caption{Human evaluation results. ``Preferred'' includes cases where annotators choose multiple systems as the best. `Approp.' is short for Appropriate, `Know.' is short for Knowledgeable.  }
    \label{tab:human_eval}
\end{table}

\begin{table}[t]
    \centering
    \small
    \begin{tabular}{l|c c c}
    \toprule 
        Model & R1 & R2 &RL \\
        \midrule 
 GPT2 + Ranking  & 19.95 & 4.70 & 16.33 \\
      GPT2 + Graph Paths  & 19.83 & 4.89 & 16.37 \\
      GPT2 + Graph Selection & 20.43 & 5.31 & 16.97 \\
      \midrule 
      GPT2 + Gold knowledge & 30.53 & 11.94 & 25.61 \\
      \bottomrule 
    \end{tabular}
    \caption{End-to-end results (in \%) on the unseen split of WoW using first turn retrieved passages as background knowledge.}
    \label{tab:exp_wow_unseen}
\end{table}

\subsection{Main Results}
We show our knowledge selection results in Table~\ref{tab:ks_holle} and \ref{tab:ks_wow}, and end-to-end  results in Table~\ref{tab:exp_holle} and \ref{tab:exp_wow_unseen}. 

From Table \ref{tab:ks_holle} we can see that our document semantic graph is helpful for the knowledge selection task and our edge-aware graph attention network is more effective in utilizing the graph structure compared to simply enumerating graph paths. 
In particular, when the graph is used, there is a large improvement in MRR when multiple gold-standard references are provided, showing that in cases where the top 1 result does not match the reference, we are able to rank the gold-standard knowledge at a high position. 

For the end-to-end evaluation in Table \ref{tab:exp_holle}, our model stands favorably among previous published results, with improvements in both knowledge selection accuracy and response quality. 

We report human evaluation results in Table \ref{tab:human_eval}. Our system scores the best in all aspects and is voted by annotators as the most preferred response in the majority of the cases.

On the WoW dataset (Table \ref{tab:ks_wow} and Table \ref{tab:exp_wow_unseen}), the basic ranking model performs slightly better on the seen split and our graph-based knowledge selection method shows benefits for generalizing to unseen topics.

\begin{table}[t]
    \centering
    \resizebox{\linewidth}{!}{ 
    \begin{tabular}{l|c c c c c}
    \toprule 
       Model  & Acc(\%)  &  MAP  & Concept & Concept  \\
       & & & MAP & MRR \\
     \midrule
   Full & 37.7 & 0.513  & 0.420 & 0.495 \\
   \midrule 
Sent. graph  & 35.6 & 0.494  & - & -  \\
Coref. graph & 37.0 &  0.510  & 0.420 & 0.421 \\
    Homog. graph & 37.3 & 0.516 & 0.409 & 0.398 \\
   \midrule 
    Sent. loss & 36.0 & 0.500 & 0.063 & 0.151 \\
    \bottomrule 
    \end{tabular}}
    \caption{Model ablations for knowledge selection on Holl-E using single reference.}
    \label{tab:ks_ablation}
\end{table}

\subsection{Analysis}

\paragraph{Model Ablations.}
We investigate whether our design of the document semantic graph is effective by exploring different variants of the document graph, including: (1) \textbf{sentence graph} with only sentence nodes and source nodes, (2) \textbf{coreference graph} that removes all AMR role edges, and (3) \textbf{homogeneous graph} that treats all edges and nodes as the same type.
The results are presented in Table~\ref{tab:ks_ablation}.
In particular, the \textbf{sentence graph}  does not make use of AMR parsing nor coreference resolution, so it only reflects the document structure. This makes it the least effective in knowledge selection and unable to perform concept selection at all.
The \textbf{coreference graph} does not perform as well as the full graph, but largely closes the gap. This suggests that entity recognition and coreference resolution are essential to the effectiveness of the document graph.
When using the \textbf{homogeneous graph}, our edge-aware graph attention network falls back to a regular graph attention network.
We can see that without edge and node semantics,  both sentence  and concept selection are negatively impacted. 

An important characteristic of our model is that it is trained to perform joint sentence selection and concept selection through a multi-task objective.
We compare our full model with a variant, which is only trained with sentence-level supervision signal. Our results show that adding the concept selection loss not only enables concept-level knowledge selection, but also improves sentence-level knowledge selection.

\paragraph{Case Studies.} 
We present some examples of the generated responses on HollE in Table \ref{tab:case_study}.
In the first example, the system started out with a comment on the character ``Morpheus'', the user agreed, and then shifted the topic towards a general comment on the movie. Both our model and the ranking model are able to follow the user's topic and make comments on the movie while the MIKe model continues the previously initiated topic.
In the second example, we see that our model and the ranking model both capture the ``viral fame'' keyword in the user's response, but our model is able to produce a more appropriate response instead of directly copying the plot.
In the last example, the ranking model repeats what the user said while our model and MIKe pick knowledge that is relevant to the rating of the movie. In this case, our model produces a more engaging response.

\begin{table*}[th]
    \centering
    \footnotesize
    \begin{tabular}{l|m{14em} | m{14em} | m{14em} }
    \toprule 
    Topic & The Matrix  & Bad Ass & Rocky V \\
    \midrule 
        History &  \textbf{S}: My favorite character was Morpheus, because he truly believed in what he was doing.
        \textbf{U}: Yeah. It was a story of a reluctant Christ-like person ...
        The Matrix ...
        should be the benchmark for all sci-fi films to come.
        & 
        \textbf{S}: What was your favorite scene in this movie?
  \textbf{U}: I liked the one in which Frank fights the skinheads.
        \textbf{S}: Yes. They messed with the wrong senior citizen!
       \textbf{U}: No doubt! Gotta love the viral fame angle they had as well. 
       & 
       \textbf{U}: My favorite character was Adrian ...
  \textbf{S}: She sure did protect her family in this great movie. 
  \textbf{U}: I have watched the rocky series probably 100 times each ... I would feel offended that Rocky V is rated lower than Rockys III and IV.\\
        \midrule 
         Ours & And so much interpretation in these movies. Have the directors ever come out and just explained it? That would make things a lot easier. & Oh yeah, it goes viral \& he is proclaimed by the public as ``Bad Ass.'' 
         & Yeah. This movie is a 10, but I feel I need to give it that bit extra due to the very low rating this movie has polled on this site. \\
         Ranking & Yeah. With The Matrix, both elements are perfectly en sync. & Right. It wasn't all that bad when Frank becomes an instant celebrity, and he and Mrs Vega (Tonita Castro) are interviewed at the morning programme LA Mornings by a journalist (Sam Rubin).
         & My favorite character is Adrian too. \\
         MIKe & I agree , and I loved the scene where he throws Neo into the subway tracks then drops down there. &  Danny Trejo's Grand Torino.
         & A wonderful movie about father and son. \\
         \bottomrule 
    \end{tabular}
    \caption{Generated responses from our system and baselines on HollE. \textbf{S} stands for system turn and \textbf{U} for user turn. 
    }
    \label{tab:case_study}
\end{table*}

\begin{figure*}
    \centering
    \includegraphics[width=0.8\linewidth]{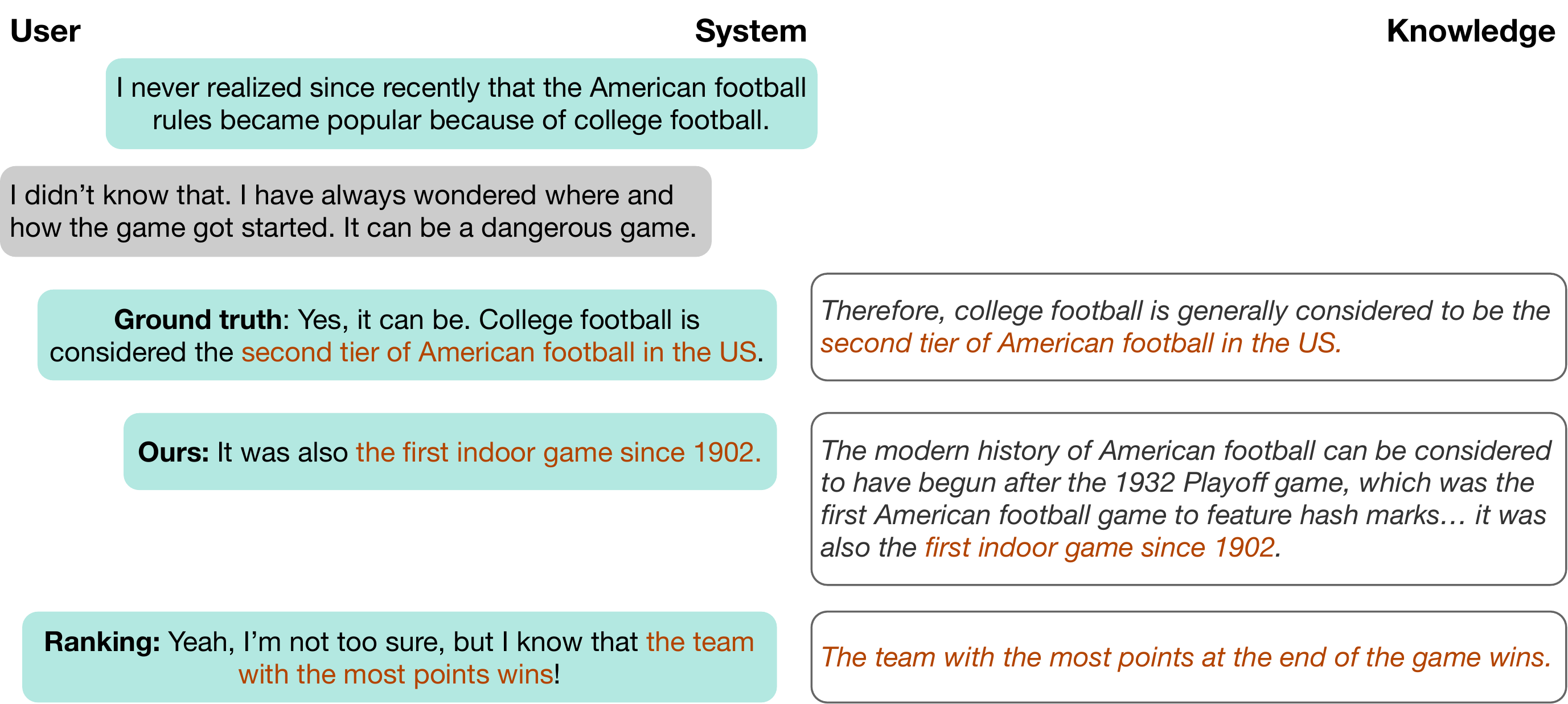}
    \caption{An example of selected knowledge and generated responses from our model on WoW.}
    \label{fig:case_wow}
\end{figure*}

Figure \ref{fig:case_wow} visualizes an example from the WoW dataset about the topic ``Football''.
In this conversation, although the knowledge selected by our model is not the same as the ground truth, it is relevant to the user's question of ``where and how the game (of football) got started''.
The ground truth, on the other hand, follows up on the wizard's own initiated topic of ``college football''.

\paragraph{Discussions on Limitations.} 
(1) \textit{Concept selection.} Current datasets were annotated with sentence selection in mind and only provided sentence level references. This makes it hard to directly demonstrate the utility of concept selection. 
(2) \textit{Better utilization of history.}
We have used the dialog history in a primitive way by concatenating the latest turns with the candidate knowledge. This ignores earlier turns, and leads to cases of repetition of history, or contradiction of persona. 
(3) \textit{Limitation of preprocessing tools.}
Our document semantic graphs rely on AMR parsing, %
which might not be available for other languages, or not be of high quality.

\vspace{-0.05in}
\section{Related Work}

\paragraph{Knowledge Selection for Dialog.}
Knowledge selection can be tightly coupled with the response generator 
~\cite{Ghazvininejad2018AKN} or performed separately prior to response generation.
Some approaches adapted question answering models ~\cite{moghe-etal-2018-HollE,qin-etal-2019-conversing, wu-etal-2021-dialki} or summarization models~\cite{Meng2020RefNetAR} for knowledge selection. 
With a pool of knowledge candidates, knowledge selection has been commonly set up as a sentence classification or ranking problem~\cite{Dinan2019WizardOW,Lian2019PostKS, Kim2020SequentialLK,chen-etal-2020-bridging, Meng2020DukeNetAD, zhao-etal-2020-knowledge-grounded}. Some work has modeled the underlying knowledge as a latent variable~\cite{Lian2019PostKS, Kim2020SequentialLK, chen-etal-2020-bridging}.
Others have explored modeling the knowledge transition over dialog turns to improve selection accuracy~\cite{Kim2020SequentialLK, Meng2020DukeNetAD, Zheng-etal-2020-difference, Zhan2021AugmentingKC}.
In comparison, we model knowledge selection as a node selection task on the document semantic graph.

\paragraph{Graph-based Knowledge Sources.}
Knowledge graphs are popular choices for integrating knowledge into dialog systems~\cite{liu-etal-2018-knowledge,moon-etal-2019-opendialkg, Xu2020KnowledgeGG,jung-etal-2020-attnio,zhou-etal-2020-kdconv}. 
However, their applicability is limited by the coverage of both entities and relations. For example in \cite{moon-etal-2019-opendialkg}, for books and movies, the knowledge base only contains metadata such as title and genre, making it impossible to conduct conversation about the actual content.
The closest work to ours is AKGCM~\cite{liu-etal-2019-augmented-graphs}, which starts from an existing general knowledge graph and then augments the knowledge graph with unstructured text by performing entity linking on the sentences.
In comparison, our document semantic graph is created from knowledge documents and during knowledge selection we select both sentences and concept nodes.

\paragraph{Application of Document Graphs.}
Document-level AMR graphs have been used for summarization~\cite{Liu2015AMRSummarization,Dohare2018UnsupervisedSA,Hardy2018GuidedNL,Lee2021AnAO} and document generation~\cite{fung-etal-2021-infosurgeon}.
Graphs constructed using OpenIE~\cite{Banko2008OpenIE} have been applied to long-form question answering and multi-document summarization~\cite{Fan2019LocalKnowledgeGraph}.

\vspace{-0.05in}
\section{Conclusion and Future Work}
In this paper, we introduce \textit{document semantic graphs} for knowledge selection.
Compared to existing document-based knowledge selection methods that typically treat sentences independently, our automatically-constructed document semantic graphs explicitly represent the semantic connections between sentences while preserving sentence-level information. Our experiments demonstrate that our semantic graph-based approach shows advantages over various sentence selection baselines in both the knowledge selection task and the end-to-end response generation task.

\section{Ethical Considerations}
The paper focuses on improving the knowledge selection component for dialog systems.

\textit{Intended use.} 
The intended use of this grounded dialog system is to perform chit-chat with the user on topics such as books and movies. 
We also hope that our released system can help research in knowledge selection.

\textit{Bias.}
Our model is developed with the use of large pretrained language models such as RoBERTa~\cite{Liu2019RoBERTaAR} and GPT2~\cite{Radford2019LanguageMA}, both of which are trained on large scale web data that is known to contain biased or discriminatory content.
The datasets that we train on also include subjective knowledge (comments on movies) that may express the bias of the writers.

\textit{Misuse potential.}
Although our system is knowledge-grounded, the output from our system should not be treated as factual knowledge. It should also not be considered as advice for any critical decision-making.

\bibliography{anthology,custom}
\bibliographystyle{acl_natbib}

\appendix

\section{Experiment Details}
Our experiments were run on a single V100 or RTX2080 GPU. We use gradient accumulation to reach an effective batch size of 16.

On average, the semantic graph construction takes 40s per document for the AMR parsing and 30s per document for coreference resolution. All documents were constructed before running knowledge selection experiments.
Our knowledge selection model requires ~10G of GPU memory and 6 hours to finish training. Our response generation model takes 1.5 hours to finish training.

We tuned our learning rate in the range of $[3e-6, 1e-5, 3e-5, 5e-5]$ and our batch size in the range of $[4, 8, 16]$. For the EGAT model, we experimented with hidden dimensions of $[50, 100, 200]$ and layers from $[2, 3, 4]$.

\section{Extra Case Studies}
\begin{figure*}
    \centering
    \includegraphics[width=\linewidth]{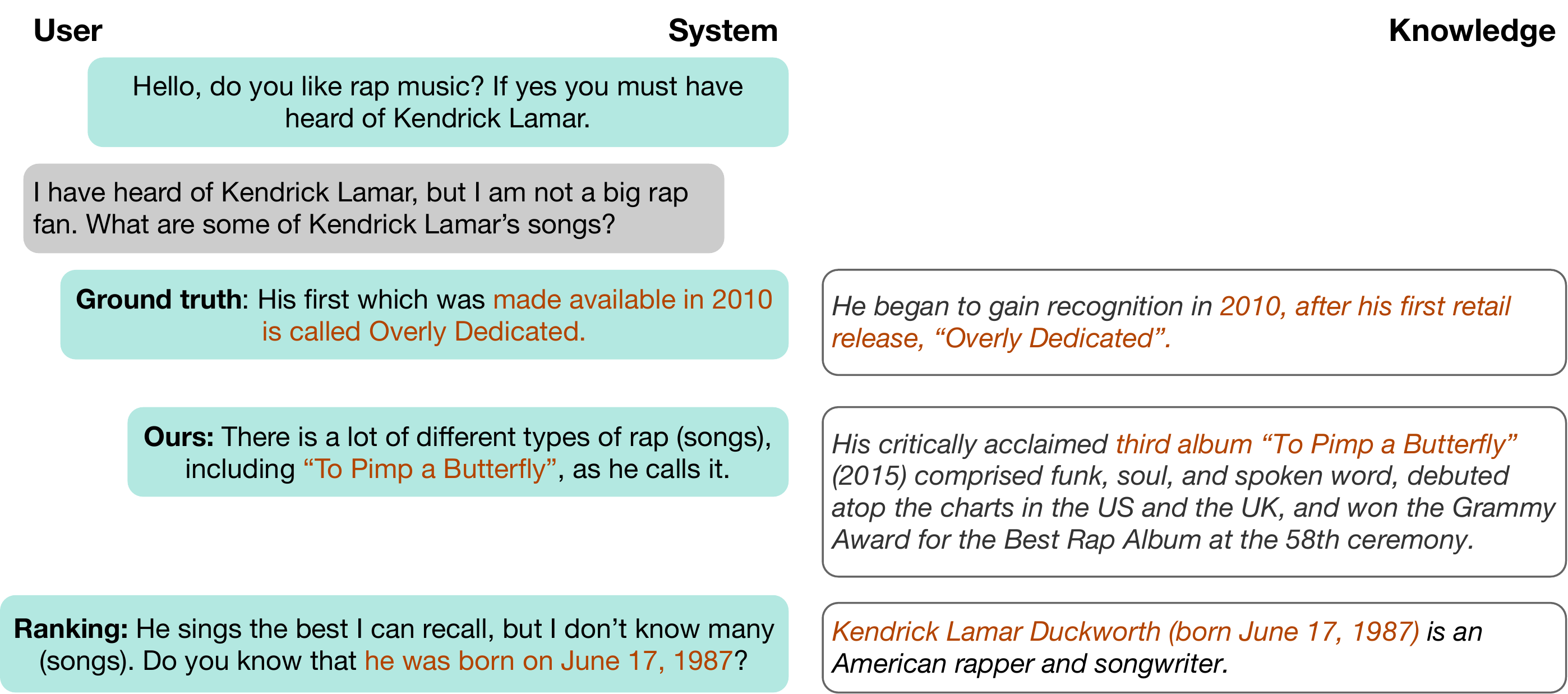}
    \caption{A case of mismatched but acceptable knowledge selection on WoW.}
    \label{fig:case_wow2}
\end{figure*}

In Figure \ref{fig:case_wow2} we present an instance where the question from the user is quite open-ended and while our model's selection does not match the ground truth, it is still relevant to the dialog and can serve as the basis for an appropriate response.

\begin{figure*}
    \centering
    \includegraphics[width=\linewidth]{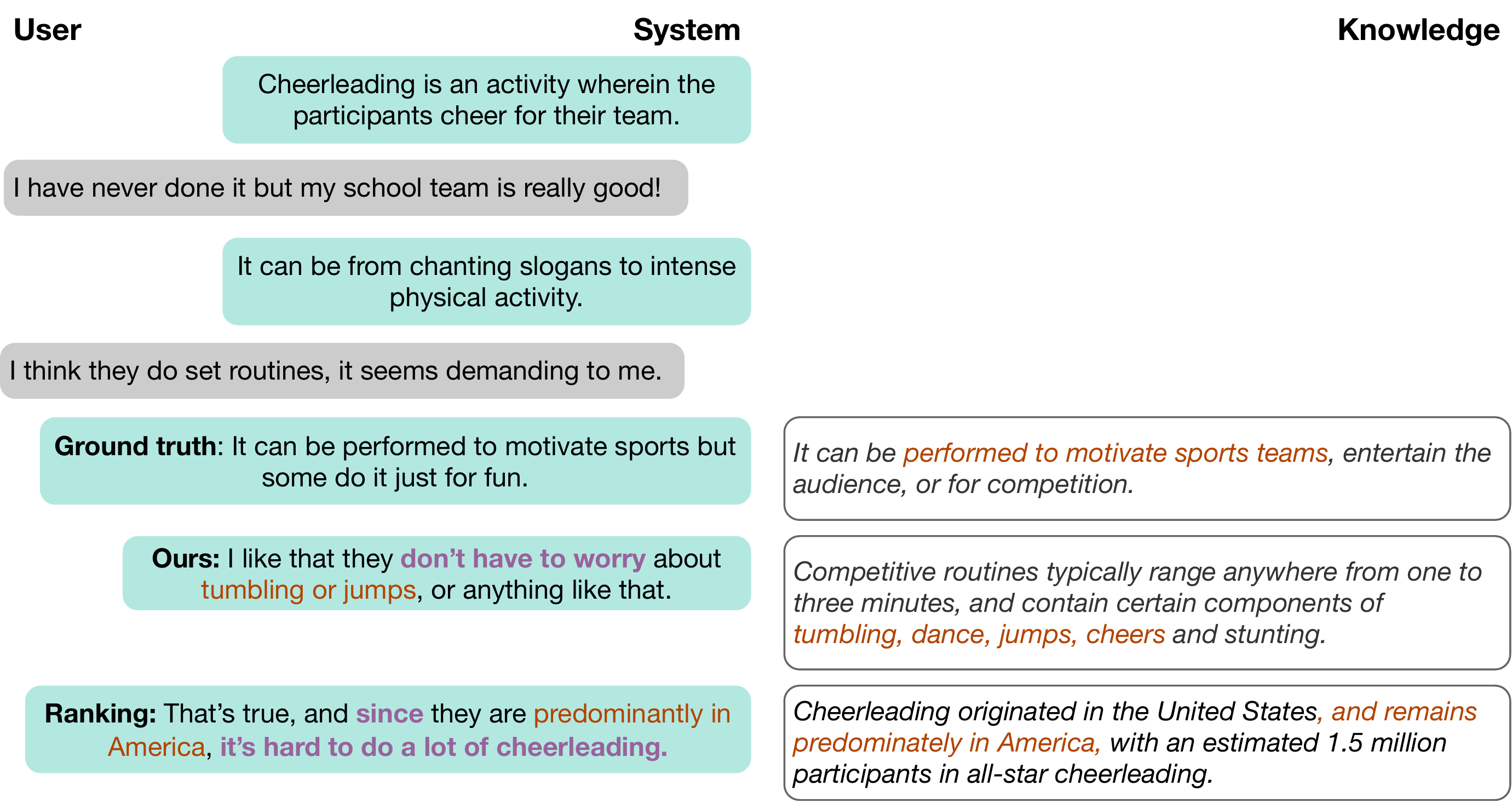}
    \caption{A case of response generation errors.The used knowledge is highlighted in brown and the generation error is marked in purple. }
    \label{fig:case_wow_error}
\end{figure*}

In Figure \ref{fig:case_wow_error} we show an example where knowledge selection performance does not directly translate to better dialog due to response generation errors.
The selected knowledge from our model follows up on the ``set routines'' mentioned by the user but the response's stance is wrong.
The baseline model selects a general statement about cheerleading as the relevant knowledge but the response is logically incorrect as the difficulty of cheerleading is not due to its geographical origin, but due to the moves.

\section{Human Eval Details}
\begin{figure*}[t]
    \centering
    \includegraphics[width=\linewidth]{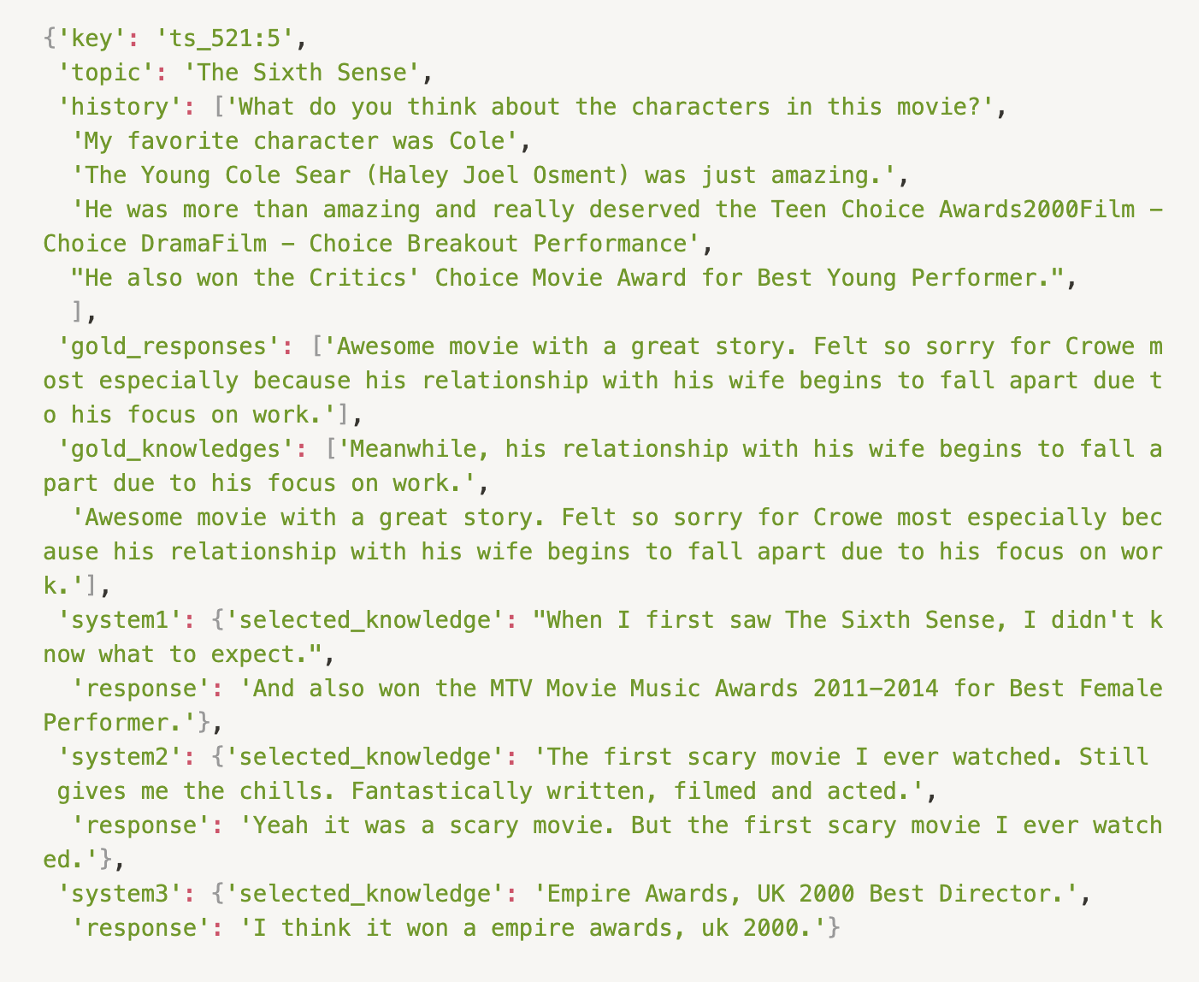}
    \caption{Example of model output provided to annotators. }
    \label{fig:annotation}
\end{figure*}

We show an example of the information provided to annotators in Figure \ref{fig:annotation}. Annotators have access to the dialog history and the ground truth responses. System outputs are anonymized.

\end{document}